\pdfoutput=1

\documentclass[11pt]{article}
\usepackage[]{naacl2021}

\usepackage{times}
\usepackage{latexsym}

\usepackage[T1]{fontenc}

\usepackage[utf8]{inputenc}

\usepackage{microtype}

\usepackage{pdfpages}
\usepackage{graphicx}

\usepackage{subfigure}
\usepackage{mathrsfs}
\usepackage{amsthm,amsmath,amssymb}
\usepackage{mathrsfs}
\usepackage{color}

\usepackage{amsmath}
\usepackage{algorithm}
\usepackage{algorithmic}
\usepackage{amsmath}
\usepackage{amsmath,amssymb}

\usepackage{booktabs}
\usepackage{multirow}

\usepackage{romannum}
\usepackage{xcolor}

\title{Imperfect also Deserves Reward: Multi-Level and Sequential Reward Modeling for Better Dialog Management}

\author{
	Zhengxu Hou$^{1}$,~ Bang Liu$^2$\thanks{~~Corresponding author.},~ Ruihui Zhao$^{1}$,~ Zijing Ou$^3$, \\
	\bf ~Yafei Liu$^{1}$, ~Xi Chen$^{1}$, ~Yefeng Zheng$^{1}$\\
	 ~~~~$^1$ Tencent Jarvis Lab \\
	$^2$ RALI \& Mila, Université de Montréal  ~~~~$^3$Sun Yat-sen University \\
	\texttt{holmes.hzx@gmail.com} ~~\texttt{bang.liu@umontreal.ca} ~~\texttt{ouzj@mail2.sysu.edu.cn}\\
	\texttt{\{zacharyzhao,davenliu,jasonxchen,yefengzheng\}@tencent.com}
}

\begin{document}

\maketitle
\pagenumbering{arabic}

\begin{abstract}
 For task-oriented dialog systems, training a Reinforcement Learning (RL) based Dialog Management module suffers from low sample efficiency and slow convergence speed due to the sparse rewards in RL.
To solve this problem, many strategies have been proposed to give proper rewards when training RL, but their rewards lack interpretability and cannot accurately estimate the distribution of state-action pairs in real dialogs. In this paper, we propose a multi-level reward modeling approach that factorizes a reward into a three-level hierarchy: domain, act, and slot. Based on inverse adversarial reinforcement learning, our designed reward model can provide more accurate and explainable reward signals for state-action pairs.
Extensive evaluations show that our approach can be applied to a wide range of reinforcement learning-based dialog systems and significantly improves both the performance and the speed of convergence.
\end{abstract}

\section{Introduction}

Task-oriented dialog systems have become a focal point in both academic and industrial research and have been playing a key role in conversational assistants such as Amazon Alexa and Apple’s Siri. \cite{budzianowski2018multiwoz,wei2018task,chen2019jddc}
Existing research on task-oriented dialog systems mainly include pipeline and end-to-end methods \cite{zhang2020recent}.
For pipeline-based systems, they usually could be divided into four components: Natural Language Understanding (NLU), Dialog State Tracking (DST), Dialog Management (DM), and Natural Language Generation (NLG).
The modular structure makes the systems more interpretable and stable than end-to-end systems, which directly take natural language context as input and output a response.

\begin{figure}[t]
    \centering
    \includegraphics[width=220pt]{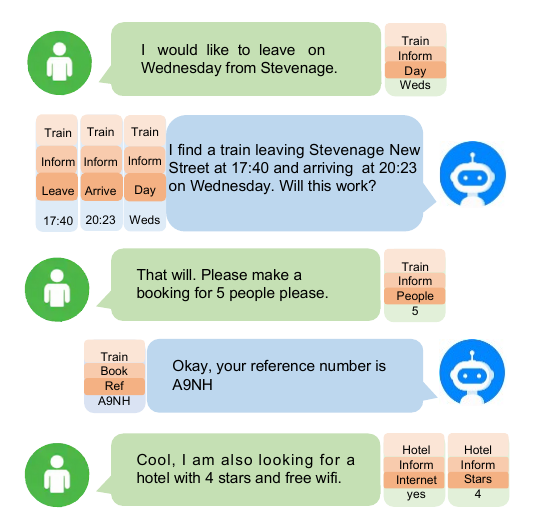}
    \caption{An example of multi-domain task-oriented dialog about train ticket booking and hotel reservation. For each utterence, we show the fine-grained actions in the form of [domain-act-slot:value].}
    \label{fig:example}
\end{figure}

In pipeline-based dialog systems, DM is a core component that is responsible for modeling the current belief state (structured information about the current situation) and deciding the next action.
Due to its non-differentiable nature, many researchers resort to Reinforcement Learning (RL) to learn a DM \cite{peng2017composite,casanueva2018feudal,chen2019agentgraph,vu2019combine}.
However, RL suffers from the problems of low data efficiency and slow convergence speed in dialog management due to reward sparsity and huge action space \cite{takanobu2019guided,liu2018adversarial,fang2019hagan}. 
To solve these problems, existing research designs reward models to estimate how similar a state-action pair is to an expert trajectory. \citet{liu2018adversarial} and \citet{takanobu2019guided} combines a Generative Adversarial Network (GAN) with RL to acquire a reward model which could give rewards in turn/dialog level. However, introducing GAN will bring other problems like model instability. To solve this, \citet{li2020guided} proposed to train a discriminator and directly use it as a fixed reward estimator during RL training.
However, there is no thorough evaluation to analyze the performance of their reward estimator. Besides, researchers ignore the huge action space of RL that has a huge impact on RL's converging speed.

In this paper, we propose to interpret a state-action pair from a multi-level and sequential perspective, instead of simply classifying them as ``right'' or ``wrong''. Fig. \ref{fig:example} shows an example of multi-domain (``Train'' and ``Hotel'') task-oriented dialog to illustrate our idea. For each utterance, we infer the domain, act, and slot of it to form a three-level hierarchy, leading to more accurate and interpretable modeling of the dialog agent's actions.
For example, for the utterance ``Okay, your reference number is A9NH'', the RL agent books a train ticket of A9NH. We infer the domain of the action ``book'' is ``train'', and the slot is ``Ref'' (reference number) with slot value ``A9NH''.
An RL agent will get rewards from the action only if it belongs to an appropriate domain, and will get slot rewards only if it takes suitable action. For example, if the RL agent in Fig.~\ref{fig:example} chooses the wrong action ``Train-Book-Ref: A9NH'' at the first turn (i.e., direct booking without confirmation from the user), it will only receive rewards for domain, since it should take the action ``inform'' instead of ``book''.

To construct a multi-level reward model, we propose the following designs.
First, we utilize a disentangled autoencoder to factorize and encode a dialog state into three independent latent sub-states, characterizing domain, act and slot, respectively.
Correspondingly, we also design an action decomposer to decompose an action into three sub-actions, taking the first user action in Figure \ref{fig:example} as an example, the decomposer will decompose action "Train-Inform-Day:Weds" into "Train", "Inform" and "Day".
Second, we learn a multi-level generator-discriminator framework. The generator generates sub-states and sub-actions from noises, and the discriminator learns to classify whether a sub state-action pair is real or generated.
In this way, the learned discriminator can give rewards to a state-action pair in terms of domain, act and slot.
Lastly, we impose Markov property to our multi-level rewards by only rewarding an act/slot if the prior domain/act is appropriate.
Such design also alleviates the problem of huge action decision space in RL, as the ``domain-act-slot'' hierarchy restricts the choice of act/slot when the domain/act has been decided.

We run extensive evaluations to test our multi-level sequential reward model by incorporating it into a variety of RL-based agents. The experimental results demonstrate that our reward model can significantly improve the performance of RL agents and accelerate the speed of convergence.



\section{Related Work}


Dialog reward modeling aims to give a proper reward to the action made by an RL agent. Traditional hand-crafted rule-based reward modeling requires expert knowledge and cannot handle unseen actions or situations. \citet{su2015reward} proposes a reward shaping method to speed up online policy learning, which models the sum of all rewards in turn level.

After that, most researchers tend to exploit GAN by considering an RL agent as a generator and a reward function as a discriminator.
\citet{liu2018adversarial} first introduce the adversarial method for reward computation. It learns a discriminator which can give the probability of authenticity in dialog level.
\citet{takanobu2019guided} further expands the adversarial method by inferring a user's goal and giving a proper reward in turn level.
However, adding adversarial training to RL will bring potential drawbacks, as training RL is different from normal GAN training whose dataset is fixed, which needs training with the environment and simulated user which is changing all the time. Thus RL and discriminator are training with a moving target rather than a fixed object. It is hard to supervise the adversarial training of generator and discriminator due to no solid feedback.
Besides, as claimed in \cite{li2020guided}, such adversarial training is only suitable for policy gradient-based methods like Proximal Policy Optimization (PPO), but not working for value-based RL algorithm like Deep Q-Network (DQN). 

Recently, \citet{li2020guided} utilizes a generator to approximate the distribution of expert state-action pairs and trains a discriminator to distinguish them from expert state-action pairs. By introducing a pretraining method, this approach can be extended to both on-policy and off-policy RL methods. However, it is still confused that whether this reward model could give correct rewards. Different from the aforementioned methods, in this paper, our model generates rewards in a more accurate sequential and multi-level manner.


\begin{figure*}[t]
    \centering
    \includegraphics[width=\linewidth]{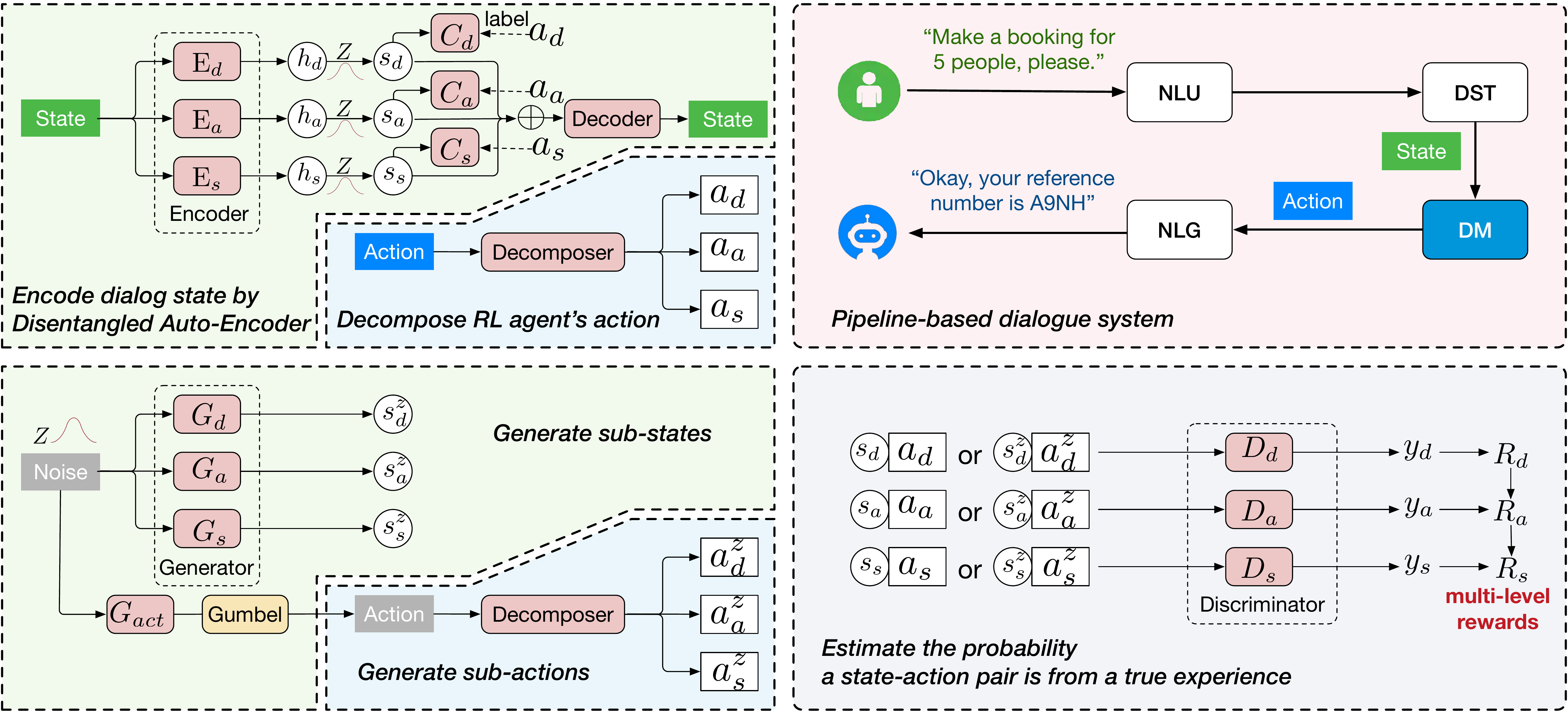}
    \caption{An overview of our multi-level and sequential reward modeling mechanism.}
    \label{fig:framework}
\end{figure*}

\section{Method}

We propose a novel multi-level sequential reward estimator and learn it under an Inverse Reinforcement Learning (IRL) framework.
IRL aims to learn a reward estimator based on expert data, which are state-action pairs $(\mathcal{S}_{e}, \mathcal{A}_{e})$ from expert policy.
IRL could be formally defined as:
\begin{equation}
\begin{split}
\mathbb{E}_{\pi_{\theta}^{*}(\mathcal{A}_{e}|\mathcal{S})}R^{*}(\mathcal{A}_{e},\mathcal{S}) \geq
\mathbb{E}_{\pi_{\theta}(\hat{\mathcal{A}}|\mathcal{S})}R^{*}(\hat{\mathcal{A}},\mathcal{S})
\end{split}
\label{reward_concept}
\end{equation}
where the goal is to find an optimal reward function $R^*$, such that based on the same states $\mathcal{S}$, expert dialog policy $\pi^{*}_{\theta}$ will obtain equal or higher rewards than the agent's policy $\pi_{\theta}$. We denote expert action and agent action as $A_{e}$ and $\hat{A}$, respectively.

Our objective is approximating $R^{*}$ by capturing the distribution of expert dialog $f_{e}$ and estimating how likely a state-action pair is from $f_{e}$ as the reward.
To accurately model the expert distribution $f_{e}$, we disentangle $f_e$ into three levels: domain distribution $f_d^{e}$, action distribution $f_a^{e}$, and slot distribution $f_s^{e}$.
Fig. \ref {fig:framework} shows the framework of our multi-level reward estimator.
Given a state-action pair from the input and output of a DM module in a pipeline-based system, we combine three components to estimate its quality.
First, we acquire sub-states and sub-actions by utilizing a Disentangled Auto-Encoder (DAE) to encode states and a rule-based decomposer to decompose actions.
Second, we learn different sub-generators to generate sub-states and sub-actions from noises.
Third, we train different sub-discriminators to classify whether a state-action pair is from expert data or agent policy. Besides, we sequentially connect the three discriminators, imposing Markov property to the multi-level rewards, as well as alleviating the problem of huge action space in RL.
Finally, the discriminators can serve as reward estimators for domian, action and slot during inference.
Algorithm \ref{algorithm__} summarizes the training process of our model components.
We introduce more details in the following.


\subsection{State-Action Decomposition and Representation}
\label{subsec:enc-dec}
We first decompose an action into sub-actions, and learn to decompose and encode a state into sub-states from domain, act, and slot level.

For action decomposition, we decompose an action $\mathcal{A}$ by rules based on how the action vector is defined. Such a rule-based decomposer can be easily implemented by first defining an assignment matrix $M$, then multiply $M$ with $\mathcal{A}$ and select three sub-spans of $\mathcal{A}$ to form three sub-actions $a_d$, $a_a$ and $a_s$, which are all one-hot vectors.

For state decomposition and representation, we decompose a discrete state $\mathcal{S}$ into sub-states of domain, act and slot, and learn a continuous representation of them by DAE.
As shown in Fig.~\ref{fig:framework}, the DAE contains three encoders $\text{E}_d$, $\text{E}_a$ and $\text{E}_s$ to extract and encode the sub-states from $\mathcal{S}$:
\begin{align}
    [h_d; h_a; h_s] = Encoder(\mathcal{S}).
\end{align}
To enforce each encoder learn the sub-state corresponding to domain, act and slot respectively, we adopt three auxilary classifiers ($C_d$, $C_a$ and $C_s$) which classify each sub-state representation ($s_d$, $s_a$ and $s_s$) with the corresponding sub-action ($a_d$, $a_a$ and $a_s$) as label. 
To enhance model generalization, we inject data-dependent noises into latent variables $h_d$, $h_a$ and $h_s$. In particular, a noise variance $\sigma^2$ is obtained via the Multilayer Perceptron (MLP) transformation from state $\mathcal{S}$: $\log \sigma^2 = MLP(\mathcal{S})$.
Then we sample noise $z$ from a Gaussian distribution $\mathcal{N}(h, \sigma^2 I)$. The reparameterization trick \cite{kingma2013auto} is further exploited to achieve end-to-end training:
\begin{align}
    [s_d; s_a; s_s] = h + \sigma \epsilon, \quad \epsilon \sim N(0, I). 
\end{align}
In this way, the sub-state representations are different for every training time of input state $\mathcal{S}$, and thus the model is provided with additional flexibility to explore various trade-offs between noise and environment.

Next, we reconstruct the state via a decoder:
\begin{align}
    \hat{\mathcal{S}} = Decoder([s_d; s_a; s_s]).
\end{align}
After that, since the state $\mathcal{S}$ is a discrete vector, we can learn DAE by minimizing a binary cross entropy loss:
\begin{align}
    \mathcal{L}_{dec} = \sum\limits_{i=1}^{|\mathcal{S}|} \left[ \mathcal{S}_i \log (\hat{\mathcal{S}_i}) + (1 - \mathcal{S}_i)\log (1 - \hat{\mathcal{S}_i}) \right].
 \end{align}
The loss for the auxilary classifiers are:
\begin{align}
    \mathcal{L}_{enc}^i =  -\frac{exp(s_i^T W_i a_i)}{\sum\limits_{a_j \in \mathcal{A}_{i}} exp(s_i^T W_i a_j)}, \ i\in [d, a, s],
\end{align}
where $W_i \in \{W_d, W_a, W_s\}$ is the learnable parameters of the classifiers and $\mathcal{A}_{i}$ is the action space of the corresponding action level. Therefore, the overall loss for training DAE is given by:
\begin{align} \label{loss_VAE}
    \mathcal{L}_{DAE} = \mathcal{L}_{dec} + \sum_{i\in{[d,a,s]}} \mathcal{L}^i_{enc}.
\end{align}

\begin{algorithm}[!t]
\caption{Reward Estimator Training}
\label{algorithm__}
\begin{algorithmic}
\REQUIRE Expert dialog $[\mathcal{S}_{e}:\mathcal{A}_{e}]$
\REPEAT
    \STATE Training DAE by Eq. \ref{loss_VAE}
\UNTIL{DAE converge}

\STATE Initialize $\mathcal{G}_{\theta}$, $\mathcal{D}_{\phi}$ with random weights $\theta$, $\phi$
\REPEAT
\FOR{g-steps}
    \STATE Sample noise samples $z$ from Gaussian prior $p(z)$
    \STATE Update $\theta$ by Eq. \ref{loss_G}
\ENDFOR
\FOR{d-steps}
    \STATE Generate $s_{d}^{z}, s_{a}^{z}, s_{s}^{z}, \hat{\mathcal{A}_z}$ by $\mathcal{G_{\theta}}(z)$
    \STATE Decompose $\hat{\mathcal{A}}$ into $a_d^z, a_a^z, a_s^z$ 
    \STATE Sample $s_{d}, s_{a}, s_{s},\mathcal{A}$ from DAE
    \STATE Decompose $\mathcal{A}$ into $a_{d},a_{a},a_{s}$
    \STATE Update ${\phi}$ by Eq. \ref{loss_D}
\ENDFOR
\UNTIL GAN converges
\end{algorithmic}
\end{algorithm}

\subsection{Adversarial Learning of State-Action Distribution}
Different from previous adversairal training methods in which generator (policy) and discriminator (reward estimator) are trained alternatively when interacting with simulated user, our GAN network \cite{goodfellow2014generative} is trained offline without the need of simulated users.
As shown in Fig.~\ref{fig:framework}, our discriminator $\mathcal{D}$ is composed of three sub-discriminators$\{ D_{{d}}, D_{{a}}, D_{{s}} \}$, our generator $\mathcal{G}$ consists a set of parallel sub-generators $\{G_{{d}}, G_{{a}}, G_{{s}}, G_{{act}} \}$ with the same Gaussian noise $Z$ as input to generate sub-states $\{s_d^z, s_a^z, s_s^z \}$ and actioin $\hat{\mathcal{A}}$.
Then $\hat{\mathcal{A}}$ is decomposed into $\{a_d^z, a_a^z, a_s^z \}$ by the same rule-based decomposer we described in Sec.~\ref{subsec:enc-dec}.
As a true action $\mathcal{A}$ is discrete, we use  Straight-Through Gumbel Softmax \cite{jang2016categorical} to approximate the sampling process.
The generators aim to approximate the distribution of expert dialog $(\mathcal{S}_{e}, \mathcal{A}_{e})$ by learning distributions $\{f_{d}^{e}, f_{a}^{e}, f_{s}^{e} \}$ of sub state-actions  with $\{G_{{d}}, G_{{a}}, G_{{s}} \}$ and $G_{act}$.
We train the generators by the following loss:
\begin{equation}
\begin{aligned}
L_\mathcal{G}(\theta)= {\mathbb{E}_{z\sim p(z)}(\log(1-\mathcal{D}(\mathcal{G}(z))))},
\end{aligned}
\label{loss_G}
\end{equation}
where $\theta$ represents the parameters of generator $G$.

For discriminator, it consists of three paralleled and independent MLP networks with a sigmoid output layer. The discriminator outputs three scores $\{ y_d, y_a, y_s\}$ that respectively denote the probability a sub state-action pair is from a true expert distribution.
The traininig loss could be written as:
\begin{equation}
\begin{split}
L_{D_i}=- [ \mathbb{E}_{(s_i, a_i) \sim f^e_i}\log D_{\phi_i}(s_i, a_i)\\
+\mathbb{E}_{z\sim p(z)}(1-\log D_{\phi_i}(s_i^z, a_i^z)) ], i\in{[d,a,s]}.
\end{split}
\label{loss_D}
\end{equation}

\subsection{Reward Shaping and Combination}
Reward shaping provides an RL agent with extra rewards in addition to the original sparse rewards $r_{ori}$ in a task to alleviate the problem of reward sparsity.
We follow the same assumption of \cite{liu2018adversarial, paek2008automating}, in which state-action pairs similar to expert data will receive higher rewards.

The rewards from our discriminators are calculated as:
\begin{equation}
\begin{aligned}
R_{d} &= y_d,\\
R_{a} &= y_a \cdot \text{Sigmoid}(\tau(R_{d} + b)),\\
R_{s} &= y_s \cdot \text{Sigmoid}(\tau(R_{a} + b)),\\
\end{aligned}
\label{rule_function}
\end{equation}
where $\{y_d, y_a, y_s\}$ are the outputs of discriminators $\{D_d, D_a, D_s\}$ in Fig.~\ref{fig:framework}.
Note that we impose Markov property into multi-level reward calculation by taking the reward of domain/act level into account when calculating the reward of act/slot level.
An agent will receive a low reward when it chooses a wrong domain even if $y_{a}$ or $y_{s}$ is high.
We accomplish this by the $sigmoid$ functions in Eq. \ref{rule_function}. $\tau$ and $b$ are two hyper-parameters controlling the shape of the $sigmoid$ function. A smaller $\tau$  will introduce a softer penalty given by prior-level reward.

After getting the three-level rewards, we propose two reward integration strategies.
The first strategy we denote as $R_{SeqPrd}$ is simply using $R_{S}$ from Eq. \ref{rule_function} as the combined reward. This strategy  will bring reward to nearly 1 or 0.
The second strategy we denote as $R_{SeqAvg}$ is computing the mean of the three rewards $\{ R_{d}$, $R_{a}$, $R_{s} \}$ as the final reward.
Finally, we augment the original reward $r_{ori}$ by adding $R_{SeqPrd}$ or $R_{SeqAvg}$ for reward shaping.



\subsection{Details of Modeling and Training}
For the Disentangled Auto-Encoder, the input of its encoder is binary states $\mathcal{S}$. We use three paralleled MLP layers with same hidden size 64 as the sub-encoders to get hidden states $\{ h_d, h_a, h_s \}$, which are the same with the architecture of the MLP network for generating noise variance $\sigma^2$.
We train the encoder, decoder, and classifier network simultaneously.

For the generator part, we utilize four independent and parallel MLP layers. All layers share the same gaussian noise as input.
The first three aim to capture the distribution in the field of $d$, $a$, $s$ with output size = 64.
The output size of $G_4$ is 300 with an output layer of ST-Gumbel Softmax.
To make the output of generators be similar with the encoding representation of DAE and bring noise to the discriminator as well, we further add two MLP networks separately after generation layer to simulate the sampling process of mean and variance.
We add weight regularization in a form of $l2$ norm to avoid overfitting.
In our experiments, the generator is weaker compared to the discriminator, therefore we set the training frequency ratio of generator and discriminator to be 5:1.

For the discriminator part, we utilize three parallel MLP layers followed by a sigmoid function as the output layer. Training a multi adversarial network is not easy. Three discriminator will be insensitive to its own field if training all $\mathcal{G}$ and $\mathcal{D}$ jointly. Thus, we train $\mathcal{G}$ and $\mathcal{D}$ in the following way.
$\mathcal{D}$ takes all outputs from $\mathcal{G}$ as input, but only chosen sub-generator and sub-discriminator pair has gradient backpropagation and others are frozen. During the experiment, we found start training from one pairs to two pairs then to all pairs brings good results.

\section{Experiments}

\subsection{Experimental Setup}
\textbf{Dataset}. We run evaluations based on the  MultiWOZ dataset {\cite{budzianowski2018multiwoz}}\footnote{https://github.com/budzianowski/multiwoz}
. It is a multi-domain dialog dataset that constructed from human dialog records, mainly ranging from restaurant booking to hotel recommending scenarios. There are 3,406 single-domain dialogs and 7,032 multi-domain dialogs in total. The average number of turns is $8.93$ and $15.39$ for single and multi-domain dialogs, respectively.

\textbf{Platform}. We implement our methods and baselines based on the Convlab platform \cite{lee2019convlab}\footnote{https://github.com/sherlock1987/SeqReward}. It is a multi-domain dialog system platform supporting end-to-end system evaluation, which integrates several RL algorithmns.


\textbf{Implementation details}.
For fair comparisons, we follow the same experiment settings in \cite{li2020guided}.
Specifically, an agenda-based user simulator {\cite{schatzmann2007agenda}} is embedded and exploited to interact with dialog agent.
We set the training environment to a ``dialog-act to dialog-act (DA-to-DA)'' level, where the agent interacts with a simulated user in a dialog act way rather than an utterance way.
We use a rule-based dialog state tracker (DST) to track $100\%$ of the user goals.
We train on millions of frames (user-system turn pairs) with $392$-dimensional state vectors as inputs and $300$-dimensional action vectors as outputs.
For all the RL networks, we use a hidden layer of $100$ dimensions and ReLU activation function.

\textbf{Evaluation metrics}.
During evaluation, the simulated user will generate a random goal first for each conversation and then complete the session successfully if the dialog agent has accomplished all user requirements. We exploit \textit{average turn}, \textit{success rate} and \textit{reward score} to evaluate the efficiency of proposed reward model. In particular, the \textit{reward score} metric is defined as
\begin{align}
\begin{split}
\textit{reward score}=\left\{
\begin{array}{rll}
-T+80,& \text{if success}\\
-T-40,& \text{if fail}\\
\end{array} \right.
\end{split}
\label{traditional_reward}
\end{align}
where $T$ denotes the number of system-user turns in each conversation session. The performances averaged over $10$ times with different random seeds are reported as the final results. Besides, we evaluate our RL models in every $1,000$ frames (system-user turn) by using $1,000$ dialogs interacting with simulated user.

\subsection{Baselines}

We evaluate the proposed reward estimator via two classical RL algorithms: \romannum{1}) Deep Q-Network (DQN) \cite{mnih2015human}, which is a value-based RL algorithm; \romannum{2}) Proximal Policy Optimization (PPO) \cite{mnih2015human}, which is a policy-based RL algorithm. 


In terms of the DQN-based methods, we compare our method $\text{DQN}_{SeqAvg}$ and $\text{DQN}_{SeqPrd}$ (corresponding to $R_{SeqAvg}$ and $R_{SeqPrd}$, respectively) with $\text{DQN}_{vanilla}$, whose reward function is defined in Eq. \ref{traditional_reward}, and $\text{DQN}_{offgan}$ \cite{li2020guided}, which also pretrains an reward function to achieve performance gains.
Similarly, we also evaluate on Warm-up DQN(WDQN) with different reward function, named $\text{WDQN}_{vanilla}$, $\text{WDQN}_{offgan}$, $\text{WDQN}_{SeqAvg}$ and $\text{WDQN}_{SeqPrd}$, respectively. 

For the implementation details of DQN-based agents, we use $\epsilon-$greedy action exploration and set a linear decay from $0.1$ in the beginning to 0.01 after $500k$ frames. We train DQN on 500 batches of size 16 every 200 frames. Besides, we use a relay buffer of size $50,000$ to stabalize training. 

In terms of the PPO-based methods, we pick up two adversarial methods: \romannum{1}) Guided Dialog Policy Learning (AIRL) \cite{takanobu2019guided}; and \romannum{2}) Generative Adversarial Imitation Learning (GAIL) \cite{ho2016generative}. AIRL works on turn level and gives reward scores based on state-action-state triple  $(s_t, a_t, s_{t+1})$. For GAIL, it works on dialog level and gives rewards after dialog ends. 

Similar to DQNs, we also compare our methods with $\text{PPO}_{vanilla}$ and $\text{PPO}_{offgan}$ \cite{li2020guided}. 
There is one extra hyperparameter named training epoch for GAIL and AIRL, which represents the training ratio of discriminator and PPO models. Here we set it to $4$. Apart from these, all the other hyperparamaters stay the same. Different from the settings for DQN, the $\epsilon-$greedy stays $0.001$ without decay. Besides, we set val-loss-coef to be $1$, meaning no discount for value loss. We also set the training frequency to be $500$ frames.

\begin{figure*}[!t]
	\centering
	\subfigure[DQN agents results]{
		\begin{minipage}[t]{0.31\linewidth}
			\centering
			\includegraphics[width=2.in]{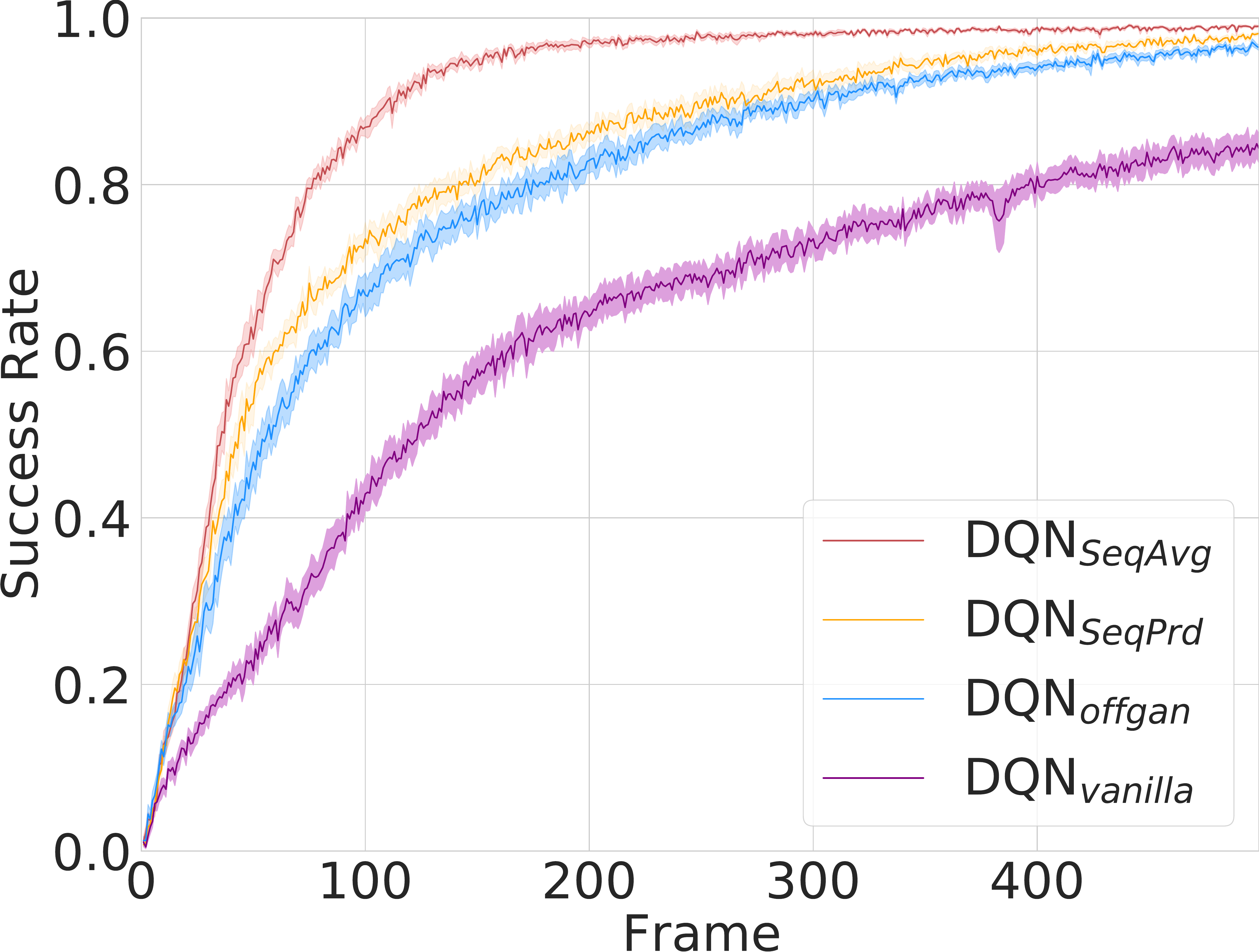} 
		\end{minipage}
	}
	\subfigure[WDQN agents results]{
		\begin{minipage}[t]{0.31\linewidth}
			\centering
			\includegraphics[width=2.in]{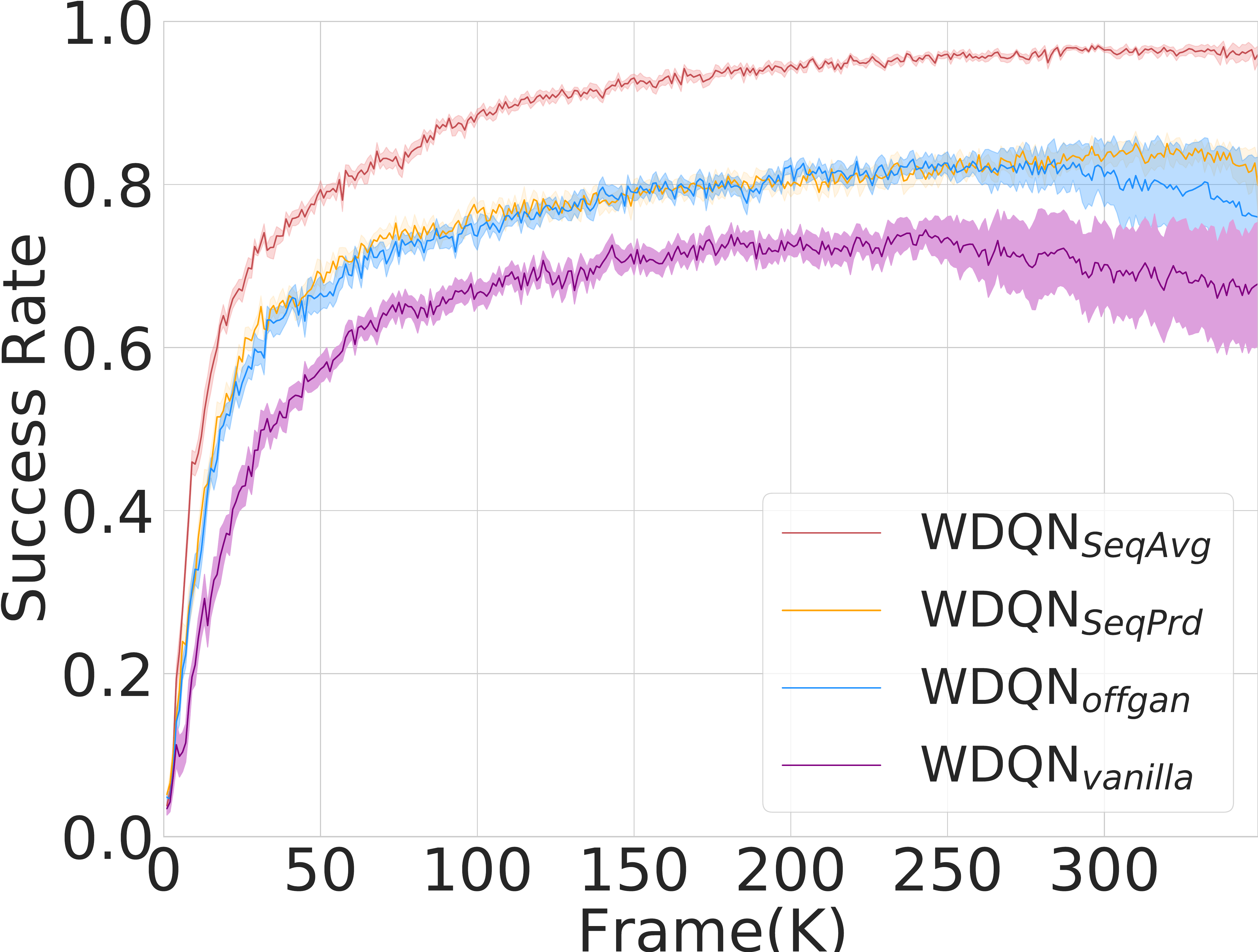}
		\end{minipage}
	}
	\subfigure[PPO agents results]{
		\begin{minipage}[t]{0.31\linewidth}
			\centering
			\includegraphics[width=2.in]{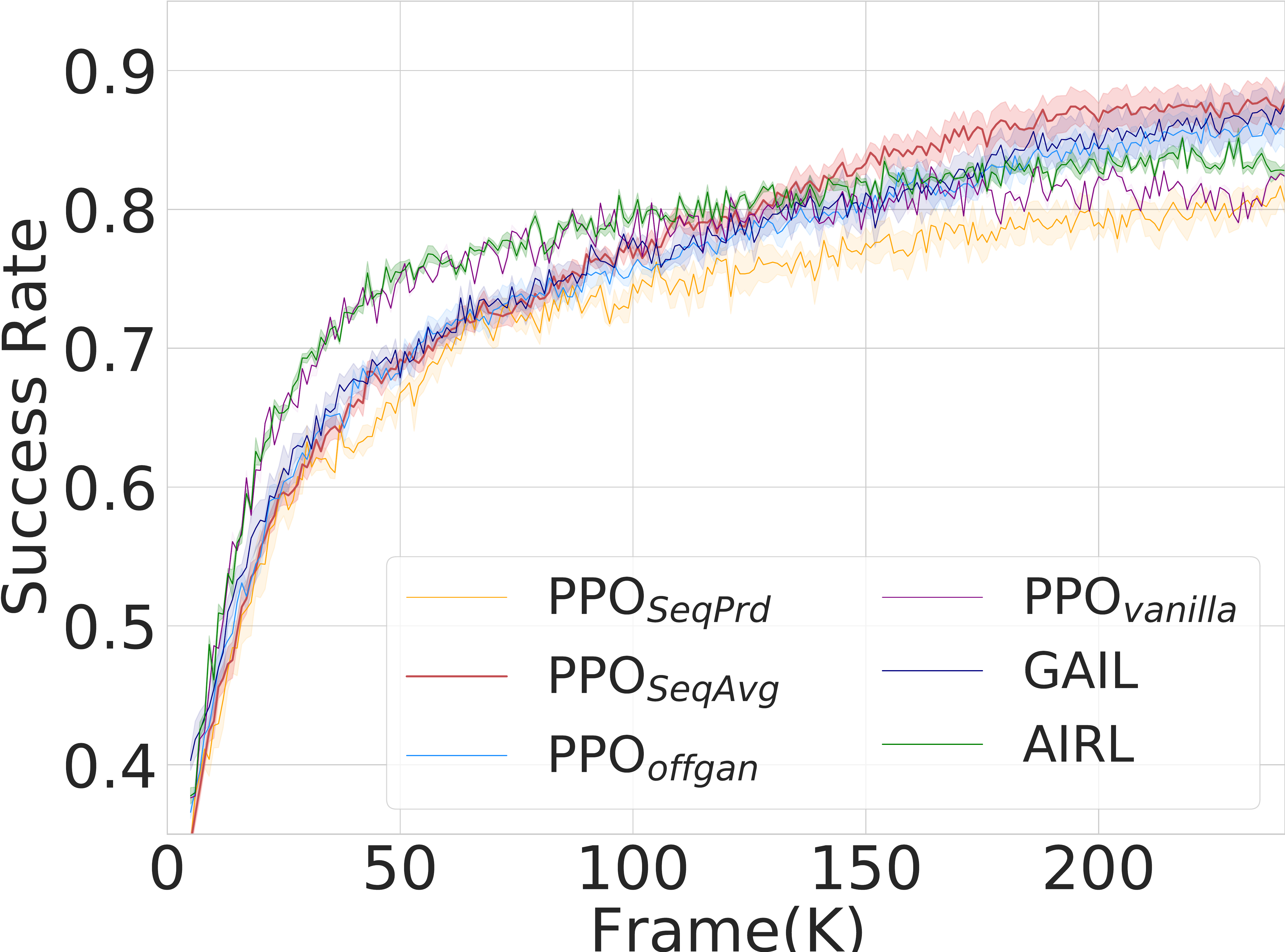}
		\end{minipage}
	}
	
	\caption{The training process of dialog agents based on different reinforcement learning algorithms.}
	\label{results_RL}
\end{figure*}

\subsection{Results with DQN-based agents}
From Fig.~\ref{results_RL} (a), $\text{DQN}_{SeqPrd}$ achieves the best performance with a success rate of 0.990 and converges after 130K, which speeds up the training process by almost 300$\%$ compared to $\text{DQN}_{vanilla}$.
Compared with $\text{DQN}_{vanilla}$, the methods using pre-trained reward functions $R_{offgan}$, $R_{SeqArg}$, $R_{SeqPrd}$ are better than vanilla in terms of both convergence speed and success rate. This phenomenon suggests that these three reward estimator could speed up dialog policy training.

Different from $\text{DQN}_{offgan}$, whose reward function is also learned by adversarial training, we further apply disentagled learning and multi-view discriminator to obtain fine-grained rewards. The performance of $\text{DQN}_{SeqPrd}$ and $\text{WDQN}_{SeqPrd}$ gains recieved in convergence speed and final performance of our methods confirm the superiorty of the hierarchical reward.

For WDQN agent, since first warmed up with human dialogs, the WDQN-based methods share similar success rate (around 6\%) before training and consistently converge faster than DQN-based models. However, the usage of warm-up operation will mislead the model to local optimum and deteriorate the final success rate. This phenomenon can be found in the last $100$ frames, the performances of $\text{WDQN}_{vanilla}$ and $\text{WDQN}_{offgan}$ drop significantly. Another attractive property of our method, compared with $\text{WDQN}_{vanilla}$ and $\text{WDQN}_{offgan}$, is the variance of success rate is obviously small, which strongly supports the remarkable benefit of exploiting disentangled representation to learn profitable sequential reward in dialog management.


\begin{figure*}[!t]
	\centering
	\subfigure[$R_{SeqAvg}$]{
		\begin{minipage}[t]{0.31\linewidth}
			\centering
			\includegraphics[width=2.in]{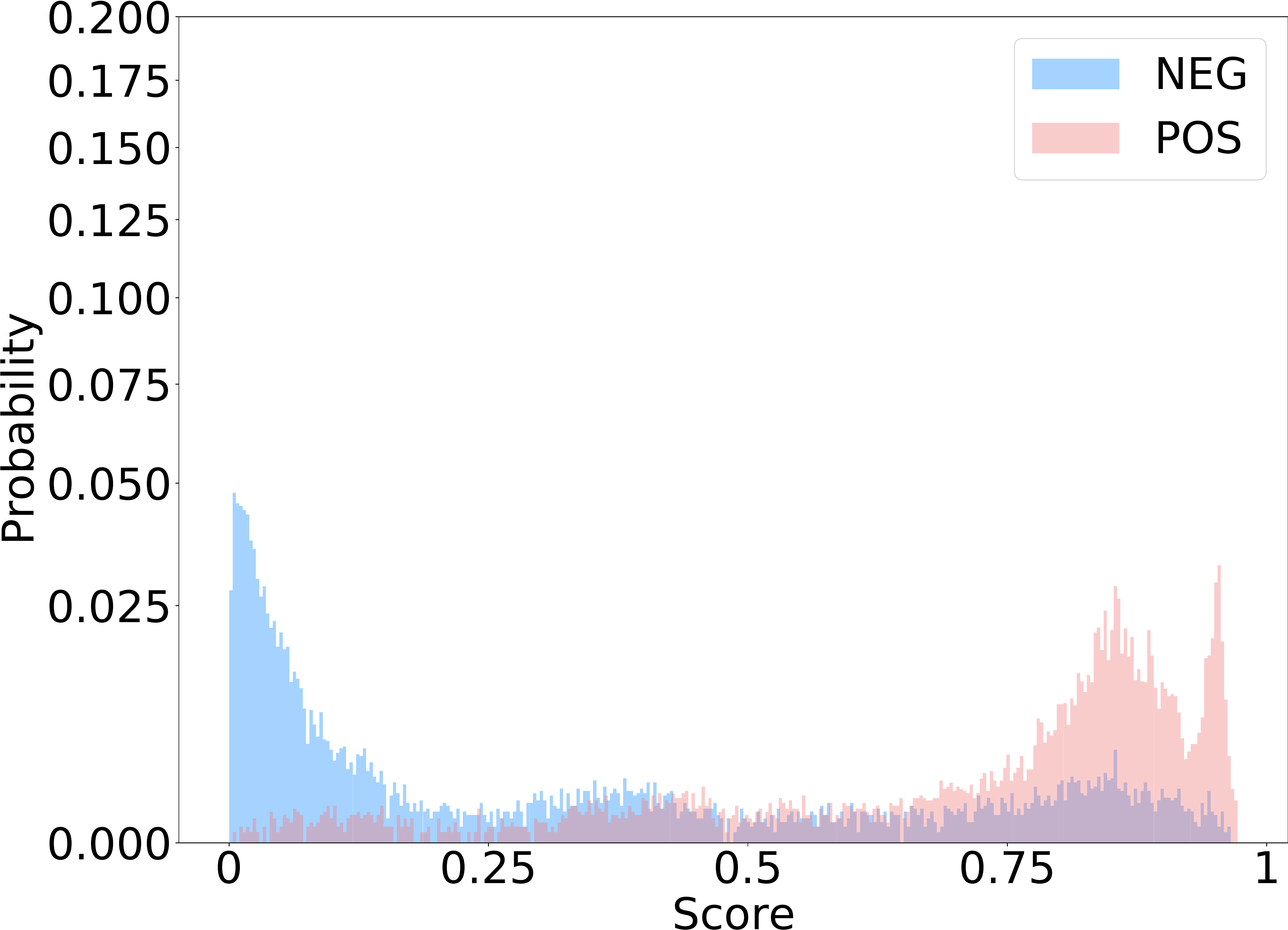} 
		\end{minipage}
	}
	\subfigure[$R_{SeqPrd}$]{
		\begin{minipage}[t]{0.31\linewidth}
			\centering
			\includegraphics[width=2.in]{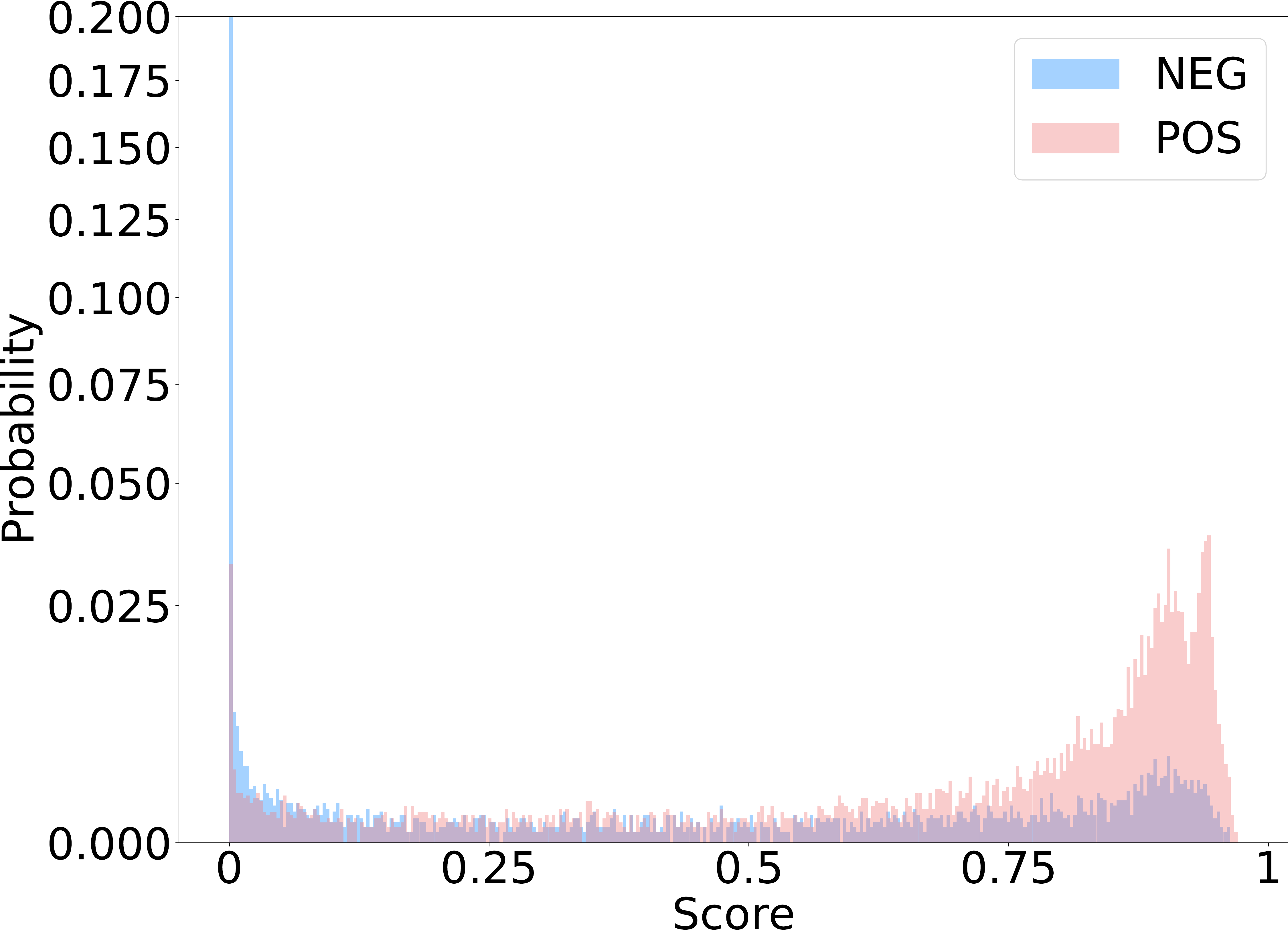}
		\end{minipage}
	}
	\subfigure[$R_{offgan}$]{
		\begin{minipage}[t]{0.31\linewidth}
			\centering
			\includegraphics[width=2.in]{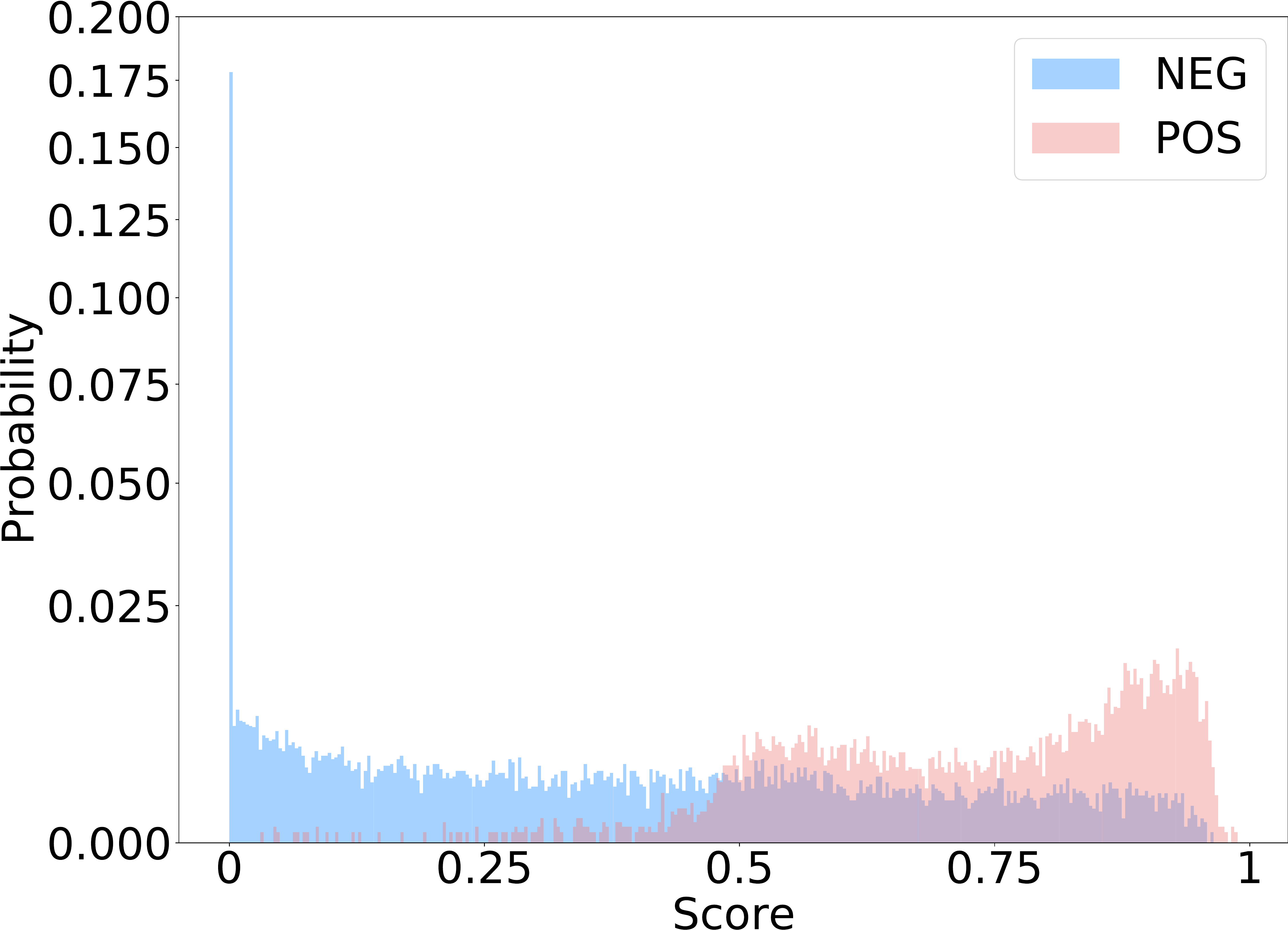}
		\end{minipage}
	}
	
	\caption{Histogram of distribution about fake/real state-action pairs in different rewards setting. Horizontal axis means the reward score, and vertical axis means frequency.}
	\label{distribution}
\end{figure*}

\begin{table}[!t]
\begin{center}
\setlength{\tabcolsep}{2.00mm}{
\small
    \begin{tabular}{lccc}
    \toprule \bf Model & SRate  & RScore & ATurn \\ \midrule
    DQN$_{vanilla}$ & 0.843 & 55.18 & 8.02 \\
    DQN$_{offgan}$  & 0.964 & 71.87 & 5.86 \\
    DQN$_{SeqAvg}$ & 0.981  &  74.21 &  5.57 \\
    DQN$_{SeqPrd}$ & \bf 0.990 & \bf 75.43 & \bf 5.40 \\
    \hline
    WDQN$_{vanilla}$ & 0.678 & 32.53 & 10.84 \\
    WDQN$_{offgan}$ & 0.760 & 43.88  & 9.33 \\
    WDQN$_{SeqAvg}$ & 0.798  & 40.01  & 8.79 \\
    WDQN$_{SeqPrd}$ & \bf 0.960 & \bf71.05  & \bf6.09 \\
    \hline
    AIRL & 0.793 & 50.63 & 10.14 \\
    GAIL & 0.832 & 51.65 & 10.01 \\
    PPO$_{vanilla}$ & 0.861 & 56.93 & 8.30 \\
    PPO$_{offgan}$ & 0.860 & 56.96 & 8.20 \\
    PPO$_{SeqPrd}$ & 0.806 & 49.86 & 9.34 \\
    PPO$_{SeqAvg}$ & \bf0.879 & \bf59.04 & \bf8.07 \\
    \bottomrule
    \end{tabular}}
\end{center}
\caption{The performance of dialog agents with different reward functions. SRate, Rscore and ATurn represent \textit{success rate}, \textit{reward score} and \textit{average turn}.}
\label{font-table}
\end{table}

\subsection{Results with PPO-based agents}
For the policy gradient based agents, we compare our models with two other strong baselines, {\it i.e.,} GAIL and AIRL, whose reward functions are updated during RL training. Similar to DQN-based methods, we employ PPO algorithms to train dialog agents with different reward functions. Before training a PPO agent, we perform imitation learning with human dialogs to warm-up PPO agents, achieving around $33\%$ success rate. For fair comparisons, we also pretrain the discriminator in GAIL and reward model in AIRL by feeding positive samples and negative samples from pretrain process of dialog agents.

As demonstrated in Fig. \ref{results_RL} (c), although AIRL rises faster than others during the first 50 frames, it converges to a worse result, compared with $\text{PPO}_{SeqAvg}$. An interesting observation is that $\text{PPO}_{vanilla}$ even performs better than AIRL. This may be due to the fact that adversarial learning is extremely unstable in RL. Therefore, we aim to learn an off-line reward function to guide the evolution of agents, as we motivate in the introduction. In the comparison between $\text{PPO}_{offgan}$ and $\text{PPO}_{SeqAvg}$, the performance gains obtained by our model verifies the advantage of exploiting multi-level reward signals. Moreover, it can be seen that, in the PPO-based RL algorithm, the performance of the agent with the reward function $R_{SeqPrd} $ is worse than that of $ R_{SeqAvg} $, but the opposite is true in the DQN and WDQN-based methods. This may be caused by that the multiplicative reward ({\it i.e.,} $R_{SeqPrd}$) may cause the gradient to be very steep, which makes the training of the policy gradient-based model unstable. However, in the value-based RL method, an average reward ({\it i.e.,} $R_{SeqAvg}$) might degenerate the performance, as a hierarchical reward is more general and intuitive, which has access to precise intermediate reward signals. The performances of the last frame in terms of \textit{success rate}, \textit{reward score} and \textit{average turn} are shown in Table \ref{font-table}, in which we could claim again that our method $\text{PPO}_{SeqAvg}$ outperforms all baseline models by a substantial margin.



\begin{table}[t!]
\begin{center}
\small
\begin{tabular}{l|cllll}
\toprule \bf Model &\bf Acc & \bf \ Prec  & \bf Rec & \bf \ F1 & \bf \ JS \\ \midrule
R$_{offgan}$ &0.79 & 0.84 & 0.76 & 0.80 & 1.39  \\ \hline
R$_{d}$ &0.86 &\bf0.97 & 0.80 & 0.88 & 0.69 \\
R$_{a}$ &0.71 & 0.91 & 0.65 & 0.76 & 0.14 \\
R$_{s}$ &0.77 & 0.95 & 0.69 & 0.80 & 0.33 \\
R$_{SeqAvg}$ & \bf0.87 & 0.91 & 0.85 & \bf 0.88 & 1.00 \\ 
R$_{SeqPrd}$ &0.87 & 0.87 & \bf0.87 & 0.87 & \bf3.73  \\
\bottomrule
\end{tabular}
\end{center}
\vspace{-3mm}
\caption{\label{compare_table} The accuracy, precision, recall, F1 and JS divergence scores on test dataset with equal number of positive and negative samples.}
\label{ablation_tabel}
\vspace{-3mm}
\end{table}

\subsection{Analysis of Different Rewards}
To visualize the model performance and what benefits a sequential reward will bring, we view the evaluation as a binary classification and distribution distance problem. we use accuracy, precision, recall and F1 to find out how good this binary model is, and use JS divergence to evaluate the ability of reward model to divide positive and negative distributions, the larger the better.
We construct a test dataset with equal numbers ($7,372$) of positive and negative samples from the test dataset. All positive samples are original state-action pairs. For negative samples, we fix states and randomly pick actions from those with different domains. We evaluate three reward models separately in Table \ref{compare_table}. $R_d$ is the best one among the three with the highest five scores. This is pretty straight forward since domain is the first identity to divide action space as groups. And for $R_a$ and $R_s$, the JS divergence is lower, this is because some actions could have different domains with the same action-slot. For example, action “Train-Inform-Arrive” and “Hotel-Inform-Arrive” have the same action-slot with different domains. Thus, $R_{a}$ and $R_{s}$ will only give an ambiguous decision boundary. But from a sequential view, we make a new combination of $R_{SeqAvg}$ and $R_{SeqPrd}$, which gives good results.

$R_{offgan}$ could give right rewards to some extent, but from Fig.  \ref{distribution} (c), there is a large intersection between fake and real distributions among three, which means it wrongly classifies fake action as right. And this is the reason why its F1 score is lower. Besides, this reward model is a biased model, its ratio of true negative and true positive samples is 0.89 thus it tends to give fake results. For both of our model, there is little bias, $R_{SeqPrd}$ is 0.99 and $R_{SeqAvg}$ is 0.98, which benefits from sequential combination.

For $R_{SeqPrd}$, $R_{SeqAvg}$ and $R_{offgan}$, $R_{SeqPrd}$ perform the best, no matter from the view of binary classification or JS divergence.
And the distribution is much sharper than $R_{SeqAvg}$ with prediction score centering at 0 or 1. For $R_{SeqAvg}$, the distribution is softer than $R_{SeqPrd}$ as shown in Fig. \ref{distribution}. Although there is no exact evaluation to say how bad one action is, but from the good results of $\text{PPO}_{SeqAvg}$, nearly the same binary classification score with $R_{SeqPrd}$ as well as lower JS divergence, we could get the conclusion that it is the most accurate rewards among the three.

    

\section{Conclusion}
We propose a multi-level and sequential reward modeling mechanism which models expert state-action pairs in terms of domain, act and slot.
Our approach combines a disentangled auto-encoder and a generator-discriminator framework to model the distribution of expert state-action pairs. The learned discriminators can thereby serve as a multi-level reward estimators.
Exierimental results show that our three-level modeling mechanism gives more accurate and explainable reward estimations and significantly boost the performance of a variety of RL-based dialog agents, as well as accelerating the covergence speed of training. 

\section{Acknowledgements}
This work was supported by the Tencent Jarvis Lab. We thank Ziming Li, Zhanpeng Huang for the helpful discussions and insightful comments.

\newpage
\bibliography{anthology}
\bibliographystyle{acl_natbib}





\end{document}